\documentclass[11pt]{article}

\usepackage{amsmath, amssymb}
\usepackage{graphicx}
\usepackage{authblk}  
\usepackage{geometry}
\usepackage[T1]{fontenc}
\usepackage{mathptmx}
\usepackage{microtype}
\usepackage[authoryear]{natbib}
\usepackage{tabularx}
\usepackage{tipa}
\usepackage{hyperref}
\hypersetup{
  pdftitle={Your Paper Title},
  pdfauthor={Author One, Author Two},
  pdfkeywords={keyword1, keyword2, keyword3},
  colorlinks=true,
  linkcolor=blue,
  citecolor=blue,
  urlcolor=blue
}

\title{Testing the Limits of Machine Translation from One Book}
\author[]{Jonathan Shaw\thanks{\texttt{jonathanshaw@xriglobal.ai}}\quad
  Dillon Mee\thanks{\texttt{dillonmee@xriglobal.ai}} \quad
  Timothy Khouw\thanks{\texttt{timothy9002@gmail.com}} \quad
  Zackary Leech\thanks{\texttt{zackleech@xriglobal.ai}} \quad
  Daniel Wilson\thanks{\texttt{danielwilson@xriglobal.ai}}}
  
\affil[]{XRI Global}
\date{\today}

\begin{document}

\maketitle

\begin{abstract}

Current state-of-the-art models demonstrate capacity to leverage in-context learning to translate into previously unseen language contexts. \cite{mtob} utilize language materials (e.g. a grammar) to improve translation quality for Kalamang using large language models (LLMs). We focus on Kanuri, a language that, despite having substantial speaker population, has minimal digital resources. We design two datasets for evaluation: one focused on health and humanitarian terms, and another containing generalized terminology, investigating how domain-specific tasks impact LLM translation quality.

By providing different combinations of language resources (grammar, dictionary, and parallel sentences), we measure LLM translation effectiveness, comparing results to native speaker translations and human linguist performance. We evaluate using both automatic metrics and native speaker assessments of fluency and accuracy.

Results demonstrate that parallel sentences remain the most effective data source, outperforming other methods in human evaluations and automatic metrics. While incorporating grammar improves over zero-shot translation, it fails as an effective standalone data source. Human evaluations reveal that LLMs achieve accuracy (meaning) more effectively than fluency (grammaticality).

These findings suggest LLM translation evaluation benefits from multidimensional assessment beyond simple accuracy metrics, and that grammar alone, without parallel sentences, does not provide sufficient context for effective domain-specific translation.
 
\end{abstract}

\section{Introduction}

Since the introduction of transformers \cite{Transformers}, neural machine translation (NMT) systems have grown in both precision and the number of languages that they support.  For example, Meta released the No Language Left Behind model \cite{nllb}, which can translate between 200 languages, followed by Google's release of MADLAD-400 \cite{madlad}, which can translate between 450 languages using their largest model.  However, achieving reasonable accuracy, especially on such a wide range of languages, requires tremendous amounts of data.  MADLAD-400 was trained on around 3T tokens.  There are more than 7000 languages on Earth, of which the vast majority lack sufficient corpora for training NMTs. On the other hand, many languages that lack corpora for training contain at least a grammar sketch, with 1904 languages having a detailed grammar of at least 300 pages \cite{glottolog}.

In light of this, placing a grammar in the context window of a large language model for translation tasks has great appeal. \cite{mtob} pioneered this approach and provided a benchmark system and result for the Kalamang language, demonstrating a score of 45.8 ChrF when translating from English into Kalamang using some heuristics to account for model context window size. To gauge how good of a result this was, they also had a linguist do the same task, achieving a score of 57.0 ChrF. \cite{gemini1.5} extended this work by introducing the Gemini 1.5 Pro model with a context window of 1 million tokens, thus accommodating the entire grammar in the model context window, achieving better than human scores of 59.1 ChrF. 

Clearly, there is potential for LLMs to improve access to Machine Translation capabilities in novel ways. However, we find that much of the discussion centers on the quality of the model and the grammar, with little attention given either to the nature of the translation task or to the evaluation of the produced translation. To take the research further, we reproduce the results in the \cite{gemini1.5} paper and adapt it to serve a new low-resource language: Kanuri. For this language, we experiment with both domain-specific and generalized domain datasets, while incorporating native Kanuri translation judgments to evaluate both fluency and accuracy.

\section{Related Work}
Model evaluation remains a critical area of research. One assumption is that models have not yet reached the theoretical upper bounds of how well they can utilize text resources such as a grammar for downstream tasks. \cite{linguini} created a benchmark of linguistic puzzles and tested a number of models on it, with the intention of finding out which models are best at linguistic reasoning, and in hopes of spurring the creation of improved models.  Of the models they tested, Claude Opus was nearly twice as good as the nearest competitor, Gemini.  \cite{grammarRag} developed a benchmark for how well a system can retrieve the requested grammatical knowledge from a corpus.  \cite{GlossLM} created a multilingual glossing model for 1800 languages.

However, \cite{badAtGrammar} cast doubt on whether the models are able to use grammars effectively, demonstrating that the LLM did not appear to use the declarative grammar for translation, although it proved useful for grammatical tasks such as producing \textit{Interlinear Glossed Text}. For translation tasks, a system given only parallel sentences and a dictionary performed comparably with a system given a grammar with parallel sentences. To follow this up, \cite{BigMTOB} extended this analysis to 16 languages, utilizing GPT-4-Turbo's longer context window of 128k tokens to prompt with increasing combinations of resources, ranging from just the extracted dictionary, including the parallel sentences, or the whole grammar book with dictionary and parallel sentences. They found that in all but three cases, the model performed \textit{better} when given only the dictionary and parallel sentences from the grammar, rather than the full grammar book. 

In many cases, the primary focus is on the data provided to the model in context; however, there is some evidence that increased prompt complexity also provides improvements.  \cite{LingoLLM} found that adding glossed sentences to the prompt helped accuracy, so \cite{GrammaMT} used the glossing model from \cite{GlossLM} to provide the gloss for the source sentence to increase accuracy.  \cite {CompTra} found that if the source sentence is split into several smaller sentences, and then the translations of these sentences are recombined, the system performs better. 

Research in the area of effective translation metrics is well established, predating modern NMT tools. \cite{lommel2014mqm} demonstrate that such evaluations are best thought of as multidimensional, creating a typology of translation error categories that include accuracy, the degree to which the translation accurately represents the meaning of the source, and linguistic conventions (formerly called fluency), the degree to which the target translation accurately represents the accepted norms for use of the language. In the context of machine-translated evaluations, one underlying principle remains clear: \cite{metaHumanEval} note that automatic metrics are intended to align with human judgments. This is not a straightforward task as automatic metrics such as ChrF and BLEU rely on objective, quantitative evaluations, while human judgments are both subjective (i.e. two native speakers may not equally agree on the suitability of a single translation) and qualitative. To that end, they designed a new metric, XSTS, which focuses on meaning/accuracy, asserting that it is more objective and less variant than fluency evaluations.

\section{Background Kanuri Language}

Kanuri (ISO 639-3: knc) is a continuum of Saharan dialects spoken by approximately 9.7 million people in Central Africa, primarily in northeast Nigeria, southeastern Niger, western Chad, northern Cameroon, and a diaspora community in Sudan \cite{ethnologue}. Its historical role as a regional lingua franca has decreased over time in favor of Hausa \cite{britannica}. The language belongs to the western Saharan subphylum of the Nilo-Saharan language family. It consists of two main dialects: Manga Kanuri (kby) and Yerwa Kanuri, or Central Kanuri (knc) \cite{ethnologue}. The Kanuri translators with whom we worked in this project describe a more complicated dialectical landscape, agreeing that while Yerwa (Central) and Manga are the two majority dialects, there are several other relevant dialects: Tumari, Bilma, Kunembu, Kwayam, Bodoi Gumati and Wuje. For this study, we focus on the Yerwa dialect (Central), which serves as the de facto language of provincial identity in the Borno, Yobe, and Gombe states of Nigeria, where it is used in education with widespread general use \cite{benton}.

Yerwa Kanuri serves as the basis for the standardized romanized orthography (known as Standard Kanuri Orthography) developed by the Kanuri Research Unit and the Kanuri Language Board \cite{Cyffer}, gaining prominence as a prominent dialect used in schools, literature, religious discourse, and media broadcasts \cite{musa}. Before this standardization, Kanuri had been written using the Ajami script (a modified Arabic script) for at least 400 years, mainly in religious or court contexts \cite{KSA2004}.   

Kanuri and its dialects were not systematically studied until the early 20th century, beginning with the work of \cite{Lukas}; however, \cite{Hutchison-1981}'s "The Kanuri Language: A Reference Grammar", provides the most comprehensive analysis of the Yerwa Kanuri grammar. His work is the primary source for the grammatical information used in this paper. Hutchison analyzed Yerwa Kanuri as having a consistent subject-object-verb (SOV) word order and nominative-accusative alignment, with grammatical relations typically marked through postpositions rather than case affixes \cite{konig2008case}.

Despite its significant speaker population and historical importance, Kanuri has a limited digital footprint and minimal NLP resources. As of 2025, Kanuri is supported by Google Translate, having been added in their major language expansion of 2024 \cite{GoogleTranslateLanguages2024}. However, the language remains notably absent from many standard NLP benchmarks and datasets. Kanuri is not included in the FLORES-101 or FLORES-200 evaluation benchmarks for low-resource machine translation, which cover 101 and 200 languages, respectively, but omit Kanuri \cite{goyal2022flores}.

The limited digital resources for Kanuri include a small annotated dataset containing 3,000 sentences with personally identifiable information released through the Lacuna Fund \cite{Lacuna_DVN/CGHWZE_2024}. This dataset was developed for named entity recognition (NER) and text classification tasks but represents one of the few structured digital corpora available for the language. Kanuri is also absent from most major multilingual NLP datasets such as OSCAR, CCAligned, WikiMatrix, and ParaCrawl.

This digital scarcity positions Kanuri as an excellent candidate for MTOB research. Although the language has a standardized orthography and official status in several countries, its digital under-representation means that machine learning systems and language technologies rarely account for Kanuri speakers. This gap persists even though Kanuri is spoken by millions of people in multiple countries, highlighting the disparity in technological development that often affects African languages \cite{adebara2022towards}. Further, Yerwa Kanuri uses a standardized Latin-based orthography (Standard Kanuri Orthography) that was officially approved in 1975, facilitating easier tokenization and processing compared to languages with non-Latin scripts. This orthographic standardization provides a stable foundation for computational approaches.

\section{The Benchmark}

\subsection{Grammar, Parallel Sentences, and Dictionary}

\cite{Hutchison-1981}'s grammar provides a comprehensive analysis of the Yerwa dialect of Kanuri, consisting of a robust 373 pages.  The grammar is organized into eight main sections, each subdivided into chapters, covering Kanuri phonology, morphology, and syntax with numerous illustrative examples. Since it had not previously been digitized, we used a custom OCR model on a scan of this grammar and had it manually post-edited to catch OCR mistakes-- a process which also ensured that this grammar was not included in the training data of the LLMs we used in this project. We used this version of the grammar in its entirety for model prompting, which amounted to more than 300,000 tokens. Additionally, we extracted 2001 parallel sentences from the grammar for use in separate prompts.
The Dictionary of the Kanuri Language, \textit{Kamus T\textschwa{}lam Kanuribe}, by \cite{cyffer1990dictionary} serves as a comprehensive lexical resource for the Kanuri language. Published in 1990 in collaboration with the University of Maiduguri in Nigeria, this dictionary contains Kanuri lexical entries with English translations and linguistic information. Each entry includes the headword, grammatical information, and semantic explanations. 
    
\subsection{Dictionary-guided and Humanitarian Test datasets}
 
One key aspect of our investigation is the way in which the quality of translation is a function of not only the model and the prompting, but also the inherent qualities of the test set.  In extremely low resource contexts, it is logical to default to creating test sets based on what is available. However, this poses problems when those test sets are not representative of either natural distributions of communication within that language (language with general utility), or high-utility domain-specific forms of communication (language that meets communication needs in a specific context). The kinds of sentences found in a grammar or dictionary should certainly represent ground truth utterances within a language; however, the primary purpose of illustrative sentences within a grammar is to illustrate a particular grammatical feature or word, not to be representative of the full range of communication that happens within a language. Thus, it is imperative for researchers to consider the ways in which the test data itself might inject artificial variance into our statements of the quality of the models we test. 

 Therefore, we crafted two datasets of 1,000 sentences, and had each of them translated by native Kanuri speakers. The first dataset is dictionary-derived, but only in the sense that we used the words as stems to generate novel sentences. These novel sentences represent not only a more challenging translation task for a model but also a more authentic representation of the complexity inherent to language used in a natural context.  

 The second dataset is derived from a glossary\footnote{https://glossaries.clearglobal.org/\#}  of terms with paired English-Kanuri translations, with a focus on health and medical terminology in a humanitarian aid context. Specifically, the glossary addresses specialized vocabulary for humanitarian contexts, covering domains such as child protection, food security, mental health support, and general medical terminology. The Kanuri translations are comprehensive, offering both literal and conceptual equivalents to complex health terminology that would otherwise lack standardized translations. Furthermore, these terms were chosen precisely because of their domain-specific utility. Example sentences for these datasets are shown in Table \ref{sentenceExamples}. 

 \begin{table}[ht]
  \centering
  \small
  \texttt{
    \renewcommand{\arraystretch}{2}
    \begin{tabularx}{\textwidth}{|X|X|}
      \hline
      \textbf{Dictionary Sentences} & \textbf{Humanitarian Sentences} \\
      \hline
      Before you start using a long grass cutter, make sure to wear protective gear such as gloves and safety glasses. & Improving access to critical facilities is crucial to get rid of unsanitary conditions in rural areas. \\
      The vendor's prices were competitive, which is the difference between making a sale and losing a customer. & I have not found an effective way to use adapted communication with my non-verbal patients yet. \\
      After graduating, she decided to travel aimlessly and see where life took her. & The research team collected biodata from thousands of participants to study the genetic factors contributing to the disease. \\
      \hline
    \end{tabularx}
  }
  \caption{Examples of Dictionary and Humanitarian Sentences}
  \label{sentenceExamples}
\end{table}
 
In both cases, the glossary and the dictionary terms are placed in the context window of the model. Thus, a humanitarian term such as 'unsanitary' or 'non-verbal patients' is presented as a known dictionary term for the model's attempt at translation, striking a balance between challenging translation tasks while still equipping the model with sufficient lexical resources to theoretically succeed at the task.

\section{Research Questions}
Since \cite{badAtGrammar} and \cite{BigMTOB} cast doubt on the value of the grammar to improve translation results, we propose a more nuanced line of investigation. Rather than asking whether grammars are useful in improving translation, we consider whether grammars are useful under certain conditions. We conjecture that when the sentences being translated are similar in domain to the sentences found in the grammar, then the parallel sentences in the grammar suffice.  However, if the source sentences are from a specific domain that is not well represented in the parallel sentences, then the LLM will benefit from the grammar in order to generalize to the new domain.

We have three research questions that we wish to address.
\begin{enumerate}
\item Does the grammar provide significant value for the LLM in domain-specific tasks?  
\item To what extent does including the grammar affect the fluency of the translation and its accuracy to the meaning of the source sentence?
\item Given a grammar and dictionary, how well can a human linguist convey meaning and create natural sentences?
\end{enumerate}

\begin{figure}[ht]
    \centering
    \includegraphics[width=0.75\textwidth]{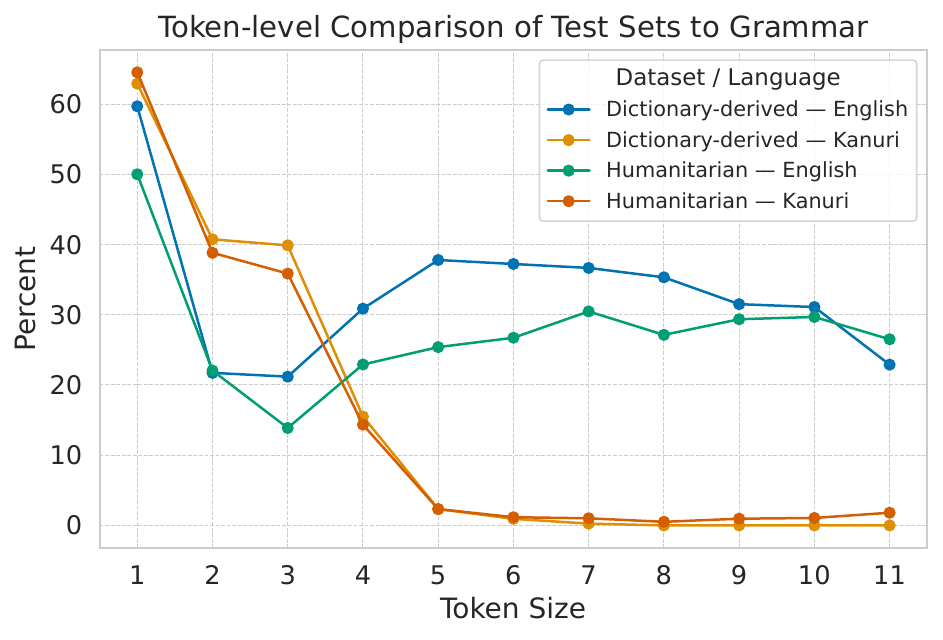}
    \caption{Token overlap with the grammar by test set and language.}
    \label{fig:ngram-overlap}
\end{figure}

To answer the first question, we compare the results of the two test sets: the dictionary-derived dataset and the humanitarian dataset. Although the construction of these two datasets is similar in that each sentence contains a particular stem from the dictionary or the humanitarian glossary, respectively, the humanitarian dataset is designed to be both domain-specific (i.e. all sentences are within the domain of humanitarian aid) and relatively distinct from the scope of the grammar, including the parallel sentences. Although it seems intuitive that a grammar would cover terms commonly found in a bilingual dictionary more than it would a glossary of humanitarian terms, we demonstrate the veracity of this intuition through a token-level analysis of coverage between the two test sets and the grammar (Figure \ref{fig:ngram-overlap}). When comparing the overlap of distinct tokens by size between the English and Kanuri sentences in the test sets and the grammar (using OpenAI's tokenizer), one finds that the dictionary dataset indeed has greater overlap with the grammar than does the humanitarian dataset.

By applying the same translation protocols consistently across the two test sets, we suggest that any meaningful difference in scores between the two test sets can be attributed to the relationship between the resources provided to the model and the test set. Thus, if the presence of the grammar positively impacts translation of the humanitarian dataset but does not affect translation of the dictionary-derived dataset, then this indicates that the value of the grammar is context dependent. These results can also be compared with results where the model is not given the whole grammar, but only the parallel sentences. 

For the second research question, we realize that ChrF scores have a high potential for false negatives in the context of low-resource languages. Neural models thrive on the observed probabilities produced in the training data, yielding models that produce not only acceptable outputs, but probable outputs. However, in low-resource contexts one can surmise that there is a 'fuzzy' zone of language expression that is less probable but still acceptable. In fact, when using a grammar as the primary source of translation, we might expect exactly that type of fuzzy output. Therefore, we hypothesize that ChrF is too coarse a metric to catch this subtlety, and thus it is possible that character-level comparison fails to acknowledge the manner and degree to which a model is, in fact, successful. To that end, we propose focusing on two aspects of translation success: fluency (how natural the translation is), and accuracy (whether the translation properly conveys the meaning).

To accomplish this, we engaged three native Kanuri speakers. These speakers provided three complementary roles for the project. First, they served as translators for the datasets themselves, yielding ground-truth data. We divided the source English sentences into chunks for them to translate, including sampled peer-validation. Their second role was for back-translation. We carefully designed protocols in which the translators would back-translate sentences that they had not seen in forward translation. Finally, their third role was as fluency evaluators, comparing paired outputs that were either peer translated, translated by the LLM, or translated by an English speaking linguist. To measure accuracy, we engaged an independent native English speaker (i.e. someone not involved in the design of the experiment) to compare the original English sentence against two round-trip translations: one that was forward translated by a native Kanuri speaker and another that was forward translated with the grammar and/or parallel sentences (LLM or linguist). Both of these forward translations were then back-translated by native Kanuri speakers.  

By comparing the forward translations at two points, we were endeavoring to isolate fluency from accuracy. In practice it is impossible to completely isolate these two elements; however, with careful prompting (to the human evaluators), one can minimize the impact of the other. When comparing the fluency of forward translations, we did not show the English sentence that acted as the source of the translation. Abstracted in such a way, there was no source to which the Kanuri sentence could be compared for accuracy, leaving fluency as the primary mode of evaluation.

When evaluating for accuracy, we showed the round-trip translations to the native English speaker. Here, it is possible that fluency will still impact this evaluation, but we contend that the variance of fluency that came from the forward translation will be reset by the act of back-translating. Additionally, we emphasized in directions to the native English speaker that the evaluation is based on accuracy (i.e. which of these two sentences best matches the \emph{meaning} of the source English sentence). 

Finally, there is only limited empirical evidence for the ability of a skilled human to perform a translation task by reference alone--especially in comparison to machine translation. Therefore, we also tasked a linguist with translating from English to Kanuri using only the grammar, parallel sentences (embedded within the grammar) and a dictionary. The linguist translator is a native English speaker and had no significant prior knowledge of Kanuri. By having the linguist translator engage in the same task as the machine translation system, we investigate the degree to which the competency of the model interacts with value of the referent resources and the demands of the task.

Here, however, we must admit that the possible answers are not equally as informative. If the linguist translator performs better than the LLM, we can infer that the LLM cannot maximize the use of grammar. If, on the other hand, the LLM is comparable to or better than the linguist translator, then we still cannot infer that the LLM is maximally utilizing the grammar. Put another way, one must consider whether the grammar itself contains the necessary information for translation success. This is a distinct and foundational question. It is of utmost importance that this question is not posed in binary terms. Instead, we must pose the following question: What degree of translation success can be achieved with a given test set using given resources and a given model for a given language? It is our hope that these findings contribute to the complex narrative of how cutting-edge technology can be successfully applied in low-resource contexts.

\section{Methods}
Our first step was to reproduce the results that \cite{gemini1.5} achieved for Kalamang. We then reproduced the experiment using OpenAI's GPT-4.1 model\footnote{https://openai.com/index/gpt-4-1/}, which has a context window of up to 1 million tokens. We found that they had general parity with GPT-4.1 achieving a score of 56.58 ChrF compared to Gemini 1.5 Pro's score of 59.00 ChrF. For the remainder of experiments, we used GPT 4.1 which we found to be the most economical due to efficient caching, while still maintaining the necessary long context window. With this in place, we took our dictionary-derived and humanitarian test sets and performed four translation protocols on subsets of both.  

\vspace{1em}
\noindent \textbf{Protocol 1: Native Kanuri speakers translations.} Interestingly, native speakers immediately identified dialect as a potential confounding variable, indicating that in some cases the derived terms were not consistent with regard to dialect--this despite the fact that both the glossary and the grammar were designed with a single focal dialect: Central/Yerwa. We directed the translators to translate the English sentence into that dialect with a 'naturalness above all' mentality. This means that if the usage of the derived term violated naturalness (e.g. the term was actually of a different dialect or the term failed to fit the particular sense expected in the source sentence), then they should feel no responsibility to keep the term and instead strive to make the sentence perfectly natural while fully capturing the meaning of the source English sentence. Even this presents some challenge; when cross-validating translation samples, native speakers would disagree on what constituted the most natural translation for a given English source sentence. This is a testament to the following factors: the complexity of the sentences, the complexity of translation as a task, the complexity of the dialectical landscape. 
    
\vspace{1em}
\noindent \textbf{Protocol 2: GPT-4.1 with dictionary and full grammar including embedded parallel sentences.} This is the same process as demonstrated in the \cite{gemini1.5} paper and entails showing the entire grammar to the LLM along with the entire dictionary and the parallel sentences as embedded in the grammar.

\vspace{1em}
\noindent \textbf{Protocol 3: GPT-4.1 with dictionary and parallel sentences extracted from the grammar.} This version further investigates the claims made by \cite{badAtGrammar}, namely that the grammar itself is not beneficial to the LLM with regard to making an effective translation. For this translation protocol we showed the LLM only the dictionary and the parallel sentences that had been stripped out of the grammar.  

\vspace{1em}
\noindent \textbf{Protocol 4: A human linguist with dictionary and full grammar including embedded parallel sentences.} For this section, the linguist translated 102 sentences from English to Kanuri following a rigorous multi-step process. First, each English source sentence was grammatically glossed, breaking it down into its constituent grammatical units. This step provided a structural blueprint for the translation process. Next, the linguist consulted a Kanuri-English dictionary to identify appropriate lexical equivalents for each grammatical unit identified in the gloss. The linguist then reviewed these preliminary translations alongside the grammatical gloss and the table of contents from \cite{Hutchison-1981} reference grammar. This review process helped determine which specific sections of the grammar would be most relevant for constructing a natural and well-formed Kanuri sentence.

The linguist then consulted these identified grammar sections to enhance the word-level translations with the appropriate morphosyntactic features of Kanuri, including case-marking postpositions, verbal tense markers, and other grammatical elements. The translations were refined to incorporate these features, paying particular attention to Kanuri's subject-object-verb word order and its system of optional postposed case markers \cite{konig2008case}. Each sentence underwent a final review by the linguist to ensure naturalness, grammatical accuracy, and adherence to the conventions of Yerwa Kanuri, resulting in a dataset of 102 translated Kanuri sentences.

\vspace{1em}
\noindent
The distribution of translation tasks by protocol and the number of comparisons of precision and fluency are described in Table \ref{tab:summary_protocols}. For the humanitarian dataset, we translated 100 sentences with the LLM with Grammar, parallel sentences, and dictionary, and another 100 with just the parallel sentences and dictionary. These 200 sentences were then paired with their correspondent native translation and evaluated twice by native speakers (who had not participated in the translation of that particular pair) for fluency. In total, we collected 400 fluency comparisons for the humanitarian dataset. Similarly for the dictionary-derived dataset we had 102 sentences translated by the LLM in both ways along with the linguist translations, paired with a total of 306 native Kanuri translations. Again, each of these comparisons was made twice, for a total of 1012 fluency comparisons across both test sets. We then back-translated these same sentences into English to produce a total of 506 accuracy comparisons (200 for the humanitarian dataset and 306 for the dictionary-derived dataset. 

\vspace{1em}
\begin{table}[htpb]
\centering
\begin{tabular}{lrrrrrrr}
\hline
Dataset & Grammar & Parallel & Linguist & Native Spkr. & Fluency & Back-trans. & Accuracy \\
\hline
\textbf{Humanitarian} & 100 & 100 & -   & 200 & 400 & 400 & 200 \\
\textbf{Dictionary}   & 102 & 102 & 102 & 306 & 612 & 612 & 306 \\
\textbf{Total}               & 202 & 202 & 102 & 506 & 1012 & 1012 & 506 \\
\hline
\end{tabular}
\caption{Experimental design: number of evaluation comparisons by dataset and translation source.}
\label{tab:summary_protocols}
\end{table}

For fluency comparisons, we asked native Kanuri speakers which of two possible sentences sounds more natural or fluent. Of the pair, one had been translated by protocols 2-4 and one had been translated by a native Kanuri speaker (protocol 1). Neither were labeled or qualified in any other way, and the source English sentence was not shown. This provides insight into the relationship between an automatic metric such as ChrF and ground-truth evaluations provided by a native speaker. What one would expect is that for any of the given protocols (other than 1), if the protocol was equally as good as protocol 1 for fluency, then the native Kanuri speaker would be equally as likely to prefer either protocol, producing a selection of 50\% from either protocol. Such results would suggest that the sources are indiscernible with regard to fluency. For human readability, we transform our results into true percentages where 100\% would represent full indiscernibility.

We then designed a similar accuracy comparison metric that was once again based on native-speaker comparisons of sentences from distinct sources. In this case, however, we had native Kanuri speakers do back-translations (being careful to ensure that speakers always translated text unrelated to their forward translations). These back-translations were then given to a native English speaker for blind comparison, in which they were shown the original source sentence and two back-translated sentences, and asked which sentence they preferred for best aligning with the meaning of the source sentence. This allowed us to explore the question: to what degree did the three non-native translation protocols accurately capture meaning as well as native speakers? Again, what one would expect is that if the forward translations captured all the meaning accurately (regardless of fluency), then it could be returned to English by means of native Kanuri back-translation in such a way that a native English speaker would be equally as likely to select from either forward source. 

We used a few methods to calculate the standard deviations. For BLEU and ChrF measurements, we used bootstrap resampling with n=1000.  To get the standard deviation of the accuracy measurements, we assumed a beta distribution with a flat prior and calculated the standard deviation of the posterior distribution given the data.  For fluency, each sentence was rated twice, once by the back translator of each variant.  Considering these samples to be independent would result in an unfairly small standard deviation, so we considered the average of these two measurements to be a sample, and calculated the standard deviation of the posterior distribution as with accuracy.
 
Averaging the ratings probably results in an overestimate of the standard deviation.  If we consider each rating to be $x$ samples, where $0.5 \le x \le 1$ and we are using $x=0.5$ and $x=1$ considers each rating as an independent sample, then as $x$ increases from 0.5 to 1, the standard deviation decreases because we have more samples.  It is unclear what the most fair value of $x$ is, but setting it to 0.5 results in the worst-case standard deviation.

With this approach, we aim to show differentiable results (i.e. fluency may not equal accuracy) and that this more nuanced approach will demonstrate more clearly which dimensions of translation are first manifested successfully, and under what conditions. 

\section{Results}

The results of these experiments are shown in Table \ref{tab:chrf_fluency_accuracy_combined}. For reference, we show Kalamang scores as run on GPT-4.1. These scores are in line with previous work \cite{mtob, gemini1.5}, and are significantly stronger than the scores we achieved in Kanuri, which trails by up to 30 points depending on the test set and the translation protocol. We then show scores for simple language instruction sentences from the \emph{Let's Speak Kanuri} handbook \cite{letsspeakkanuri}. These scores are the strongest of the Kanuri sentences, demonstrating the impact of the authenticity of the translation task. It is one thing to translate a simple declarative sentence; it is another thing to translate sentences that have a complex utility in a specific domain context.

\begin{table}[h]
\centering
\begin{tabular}{lrrrr}
\hline
\textbf{Domain / Protocol} & \textbf{ChrF (\%) ± $\sigma$} & \textbf{ChrF++ (\%) ± $\sigma$} & \textbf{Fluency (\%) ± $\sigma$} & \textbf{Accuracy (\%) ± $\sigma$} \\
\hline
Kalamang / Zero-shot            & 17.40 ± 1.03 & 16.09 ± 0.93 & -- & -- \\
Kalamang / Grammar only         & 44.68 ± 2.50 & 40.53 ± 2.57 & -- & -- \\
Kalamang / Parallel only            & 56.65 ± 3.20 & 54.19 ± 3.01 & -- & -- \\
Kalamang / Grammar + Parallel   & 58.17 ± 3.00 & 55.07 ± 2.93 & -- & -- \\
Let's Speak Kanuri / Grammar    & 27.42 ± 1.86 & 26.65 ± 1.92 & -- & -- \\
Let's Speak Kanuri / Parallel   & 32.58 ± 1.98 & 31.63 ± 1.90 & -- & -- \\
Kanuri Dictionary / Zero-shot   & 13.57 ± 0.93 & 10.36 ± 0.70 & -- & -- \\
Kanuri Dictionary / Grammar     & 17.71 ± 0.40 & 13.69 ± 0.31 & 8.65 ± 3.97 & 21.15 ± 6.00 \\
Kanuri Dictionary / Parallel    & 19.19 ± 0.46 & 14.95 ± 0.36 & 15.38 ± 5.20 & 28.85 ± 6.86 \\
Kanuri Dictionary / Linguist    & 19.29 ± 0.42 & 15.98 ± 0.36 & 11.54 ± 4.55 & 26.92 ± 6.66 \\
Kanuri Humanitarian / Zero-shot & 15.98 ± 0.40 & 12.17 ± 0.31 & -- & -- \\
Kanuri Humanitarian / Grammar   & 23.40 ± 0.71 & 18.41 ± 0.62 & 6.86 ± 3.59 & 35.29 ± 7.51 \\
Kanuri Humanitarian / Parallel  & 24.34 ± 0.64 & 19.47 ± 0.57 & 8.82 ± 4.05 & 35.29 ± 7.51 \\
\hline
\end{tabular}
\caption{ChrF and ChrF++ scores (with standard deviations from bootstrapping), along with fluency and accuracy preferences (± $\sigma$), across translation domains and protocols.}
\label{tab:chrf_fluency_accuracy_combined}
\end{table}

In line with previous research, these results demonstrate that showing the grammar or parallel sentences is superior to zero-shot prompting. Similarly, parallel sentences outperformed the full grammar prompting, being on par with the linguist. This suggests that a state-of-the-art LLM is able to perform translation tasks at a level comparable to a skilled human, but that a grammar in itself is an insufficient source for successful translation of more complex source sentences. 

Next, it should be noted that the domain-specific humanitarian dataset outperformed the dictionary-derived dataset by several points, both in ChrF and in accuracy. Here, it is important to acknowledge that the prompting treated humanitarian glossary terms (which were sometimes phrasal) as dictionary terms, which were also provided to native translators. We hypothesize that this benefited the model in two critical ways: first, the model was presented linguistic information that frequently had constituency structure embedded in it, and such phrases were more likely to survive the kind of direct translation that occurs in a naive (i.e. grammar-based translation) environment. Second, we observed that some test sentences in the humanitarian test set contained multiple words derived from the glossary. Given the domain constraints, it should come as no surprise that the glossary was sufficiently robust to contain semantically related terms to the target term. 

The difference between fluency and accuracy is marked, suggesting that neither grammar nor a set of parallel sentences suffices for an LLM to achieve fluent translations: native speakers rarely preferred the fluency of any of the three translation protocols over native translation (<15\%). Although accuracy (meaning) is also quite low, it is distinct from fluency, reaching a 35\% preference in the humanitarian domain. It is important to note that the fluency and accuracy judgments represent an absolute preference, not an absolute judgment of meaning, which is to say that these percentages do not mean that 65\% of non-native translations have no accuracy, but rather that 35\% of the time the accuracy of non-native translations is indiscernible from native translations. We maintain that these judgments ground the ChrF scores in a concrete relationship to speaker judgments in an objective way, with the added advantage of looking at translation quality from a multidimensional perspective.

To further explore the implications of these results, samples of both successes and failures by protocol for accuracy are shown in Table 4. Each protocol shows an example success (preferred AI or linguist) and an example failure (preferred native) with the original source sentence.   

\begin{table}[h]
\centering
\begin{tabular}{llp{4cm}p{4cm}p{4cm}}
\hline
Sample & Preferred & Original & AI Translation & Native Translation \\
\hline
1 & Native & While operational coordination is crucial in responding to outbreaks of infectious diseases like the common cold, it is equally important for addressing more severe health threats. & If the procedures are followed in making the virus disappear disathe [sic] sickness will be most important but that may affect and risk other health related issues & As the prevention of communcable diseases like cold is important, so also prevention of other diseases is equally important. \\
2 & AI & Moreover, it is obvious that he doesn't have the ability to cook a decent meal. & It is also obvious that she can't cook well. & After that, he explained that he cannot cook because he does not have the knowledge to cook. \\
3 & Native & If a person is shouting loudly in a public place, it's best to stay away from them. & Whoever constantly investigates where meetings are being held should always be informed. & When a person starts shouting in public, the best thing to do is to stay away. \\
4 & AI & It's a good idea to educate new mothers about Infant and Young Child Feeding (IYCF) practices to promote healthy growth and development. & It is expected that the women will be enlightened about the best food to give to children that is known as IYCF so that they will grow stronger. & It is very important to sensitise lactating mothers on child care to enable growth and good health. \\
5 & Native & The toad's ability to camouflage itself is impressive, and the fact that it can change its skin color is even more remarkable. & Some frogs are so surprising, the way they hide themselves is funny, they can camouflage in the surrounding. & The effort of the salamander to change itself is truly remarkable, and without a doubt, it can change its skin color as it pleases. \\
6 & Linguist & The tobacco flower is a common ingredient in traditional teeth coloring practices among women in some communities. & There are multiple types of bitter tomato used for teeth among some women. & Some flowers are mainly used by women to clean their teeth. \\
\hline
\end{tabular}
\caption{Examples of success and failure cases across protocols for accuracy comparisons.}
\label{tab:sample_backtranslation_examples}
\end{table}

These data illuminate some of the complexities of translation and demonstrate that in low resource contexts even native speakers can struggle to accurately capture the meaning in a round-trip translation task. For example, the word 'toad' is returned to English both as 'frog' and 'salamander'. These variations of accuracy are ubiquitous throughout just these samples and suggest that a metric such as ChrF is inadequate to capture such nuance. To be sure, in some cases, translation protocols 2-4 are significantly failing in terms of accuracy. A clear example from the samples is sample 3. The AI translation struggles both to be coherent unto itself, and to accurately capture the meaning of the original sentence. On the other hand all native-translation round trip examples--even the ones where the AI was deemed more accurate--still capture the general meaning of the original source sentence.

Additionally, in each of the samples, there are cases where a term in the original source sentence is semantically present in both the AI translation and the native translation, but in a different form. These are instances where ChrF would fail to provide a reliable signal, despite partial successes in the translation. In sample 1, the AI translation uses 'virus', while the native translation uses 'communicable', which presumably corresponds with the source term 'infectious'. Similarly in sample 6 the source term 'tobacco flower' is inappropriately returned as 'bitter tomato' for the AI translation, while the native translation references the more generic 'flower'. Perhaps the best way to consider these results is as the real-world representation of the children's game 'telephone' in which a source message is passed through multiple mediums (generally several subsequent speakers), and then the message is reported, allowing all who participated to marvel at how much the message has changed as a result of transmission and reception. However, in that game, the quality of the medium is generally considered equal among the various participants. No one is laughing because 'so-and-so really doesn't know what he's doing', but rather because the very process itself warps the signal. Here, however, we are asking 'What happens to linguistic meaning as transmitted through a medium other than a native speaker and is instead passed through a hybrid medium composed of a language model (either an LLM or a linguist) in conjunction with a grammar?' While the results are sometimes compelling, in other cases, the results are humorous at best and suggest that a grammar is not simply a condensed and codified language model waiting to be unpacked and plugged into an LLM. 

\section{Discussion}

In light of these findings, we conclude that a grammar does not add substantial value when generalizing into a new domain, such as going from a general domain to a humanitarian domain. In contrast, the grammar continues to be a stumbling block for AI translation compared to typical methods of n-shot prompting. Simply put, it seems possible that a collection of parallel sentences models a given language \textit{better} than a grammar. Of course, it is possible that the issue is not the grammar itself, but rather the compatibility of the presentation of the grammar with the recipient. However, if this were the case, one would expect that a linguist (i.e. the kind of people that frequently make grammars) would be able to leverage a grammar in a way that is significantly different from the AI model. That is not the case. Still, it is important to remember that grammars have long and well-established utility, including helping speakers better understand how a language behaves. Surprisingly, our findings suggest that such utility remains distinct from language modeling itself.

This analysis is true not just for traditional automatic metrics, but also more nuanced evaluations derived from native-speaker judgments for both accuracy and fluency. Including the grammar results in lower scores across all metrics--though here it is important to note that more data would be valuable to confirm these results, as in most cases the difference is within one standard deviation. 

Finally, in this case, the results between the linguist and the LLM remain very close. While the linguist was able to achieve stronger scores by automatic metrics (ChrF), he failed to surpass the model's ability to render fluency and accurate meaning when the model was solely attending to parallel sentences. This suggests that the LLM is able to maximize the utility present in the parallel sentences; however, its limits in getting useful knowledge out of a grammar suggest that a grammar is not a good approximation of a language when performing a translation task. As AI continues to become more advanced, it is plausible to expect that the declarative section of a grammar will be useful for the task of translation, but we believe it is equally plausible that it will continue to be harmful. The accuracy "boost" from adding the declarative section of a grammar may be negative. An optimally rational entity will likely create worse sentences with a grammar and dictionary than if it simply had parallel sentences. This, of course, comes with the caveat that not all grammars and not all parallel corpora are equal (either in size or quality). Though it would seem, as a rule, parallel corpora are a more efficient source for mining the kind of information required for building a language model. 

Having thus explored the value of the grammar, we now turn our attention to exploring two other elements, which upon inspection prove to be critical to the discussion. The first relates to the nature of the metrics--specifically ChrF derived metrics, and the second relates to the nature of source sentences. 

Regarding metrics, our research aligns with \cite{lommel2014mqm}, indicating that various dimensions of translation success do not occur at the same time and that the lower range ChrF is largely unable to parse these distinctions. It is important to remember that the only evaluations that truly matter are native-speaker evaluations. Thus, ChrF is a solid proxy for translation success at higher scores, but provides poor signal for why a translation fails in the context of translation failure. We offer our nuanced analysis of fluency and accuracy--a time-intensive task that encourages continual partnerships with native speakers--as a snapshot of the relationship between ChrF scores and real success. In resource-rich language contexts, such distinctions may prove insignificant, but in contexts such as Kanuri, where success may be partial for some time, the distinction of a few points of ChrF might have profound impacts on speaker judgments across multiple dimensions. 

Regarding the nature of source sentences for translation, we acknowledge that many scores are easily abstracted by scale. It is much easier to evaluate a model using averaged automatic metrics with little attention given to whether the test sentences are properly representative of the language. In this paper, we have attempted to pull the veil back to some degree and acknowledge that a fully robust language model is able to handle sentences of diverse difficulties across a diverse set of domains. Here, we demonstrate that translation quality is highly dependent on these two factors and suggest that careful consideration should be given to the heterogeneity and authenticity of test datasets. Simple tests yield simple results.

\section{Conclusion}
Advancements in AI continue to occur at such a pace that in many cases the creation of the tool outpaces our ability to measure the efficacy of the tool. This is certainly the case with LLM's where there is an enthusiasm for discovery that must be balanced against a level-headed consideration of the ways in which the tools for measurement characterize such discoveries. AI has proven in many contexts to be remarkably effective in modeling the translation from one language to another. However, we must avoid false positives derived from simple measurements of complex outputs, such as translation. We find that a more granular approach, both with regard to test set diversity and metric diversity yields a nuanced answer that better approximates the "success" continuum.

Based on this nuanced answer, we suggest that future research could benefit from both a greater quantity of data (both within and across languages)--as well as a greater diversity of questions relating to what exactly is being measured. For future work, we suggest constructing compound test sets, built with domain diversity and translator perplexity in mind from the ground up. Additionally, we suggest eliciting speaker judgments that relate not only to fluency and accuracy, but other dimensions that indicate translation success, and exploring the gradient relationship between ChrF scores and the various dimensions of translator quality. In this way, we hope that such research can bring much needed attention to the question of not only when language models achieve success, but also the ways in which success manifests across various dimensions at various rates. 

\section{Ethics Statement}
Our mission is founded on the desire to promote human flourishing through language access globally. We acknowledge that this task requires careful and continuous consideration, involving many people and organizations in cross-cultural and cross-lingual contexts. We endeavor to ensure that this work can benefit the individuals and language communities with which we partner, while also striving to identify and mitigate any possible risks associated with artificial intelligence or language translation. We gratefully acknowledge those who served as translators and those who provided native-speaker judgments for their contributions to this work. With their permission, we recognize them by name. We do not collect or retain any other personally identifiable information. It is our sincere hope that the focus on humanitarian terms in this paper ultimately benefits Kanuri speakers in practical ways.

\section{Acknowledgments}
We thank the following people for helping to make this publication possible (presented alphabetically).
Abubakar Mustapha Shettima for his work in translating and evaluating Kanuri sentences.
Ali Modu Kagu for his work in translating and evaluating Kanuri sentences.
Bashir Ibrahim for his work in translating and evaluating Kanuri sentences.
John Hutchison for his work in creating Kanuri language resources and graciously allowing us to use these resources.
Mustapha Alkali Kolo, Independent International Language Service Provider, for his work in translating and evaluating Kanuri sentences.
William Darren Dinan for his work in evaluating accuracy of English backtranslations.

\bibliographystyle{plainnat}
\bibliography{bib}

\begin{thebibliography}{31}
\bibitem[Adebara and Abdul-Mageed(2022)]{adebara2022towards}
Ife Adebara and Muhammad Abdul-Mageed.
\newblock Towards afrocentric nlp for african languages: Where we are and where we can go, 2022.
\end{thebibliography}

\end{document}